\documentclass[journal]{IEEEtran}

\usepackage{graphicx}
\usepackage{amsmath}
\usepackage{multirow}
\usepackage{array}
\usepackage{amssymb}

\ifCLASSINFOpdf
\else
\fi
\begin{document}
%
\title{An End-to-End Foreground-Aware Network for Person Re-Identification}
%
%
%

\author{Yiheng Liu,
        Wengang Zhou,
        Jianzhuang Liu,
        Guojun Qi,
        Qi Tian,~\IEEEmembership{Fellow,~IEEE}, \\
        Houqiang Li,~\IEEEmembership{Fellow,~IEEE}
        \thanks{Yiheng Liu, Wengang Zhou and Houqiang Li are with the CAS Key Laboratory of Technology in Geo-spatial Information Processing and Application System, Department of Electronic Engineering and Information Science, University of Science and Technology of China, Hefei, 230027, China. \protect\\ E-mail: lyh156@mail.ustc.edu.cn, \{zhwg, lihq\}@ustc.edu.cn.}
        \thanks{Jianzhuang Liu is with the Noah’s Ark Lab, Huawei Technologies Company Limited, Shenzhen 518129, China. \protect\\ E-mail: liu.jianzhuang@huawei.com}
        \thanks{Guojun Qi was with Huawei Cloud EI Product Department. \protect\\ E-mail: guojunq@gmail.com.}
        \thanks{Qi Tian is with Cloud \& AI, Huawei Technologies. \protect\\ E-mail: tian.qi1@huawei.com.}
        \thanks{Corresponding authors: Wengang Zhou and Houqiang Li.}
        }

%
%

\markboth{Journal of \LaTeX\ Class Files,~Vol.~14, No.~8, August~2015}%
{Liu \MakeLowercase{\textit{et al.}}: An End-to-End Foreground-Aware Network for Person Re-Identification}
%



\maketitle

\begin{abstract}
  Person re-identification is a crucial task of identifying pedestrians of interest across multiple surveillance camera views. For person re-identification, a pedestrian is usually represented with features extracted from a rectangular image region that inevitably contains the scene background, which incurs ambiguity to distinguish different pedestrians and degrades the accuracy.
  Thus, we propose an end-to-end foreground-aware network to discriminate against the foreground from the background by learning a soft mask for person re-identification. In our method, in addition to the pedestrian ID as supervision for the foreground, we introduce the camera ID of each pedestrian image for background modeling.  The foreground branch and the background branch are optimized collaboratively.  By presenting a target attention loss, the pedestrian features extracted from the foreground branch become more insensitive to backgrounds, which greatly reduces the negative impact of changing backgrounds on pedestrian matching across different camera views.
  Notably, in contrast to existing methods, our approach does not require an additional dataset to train a human landmark detector or a segmentation model for locating the background regions. The experimental results conducted on three challenging datasets, \emph{i.e.}, Market-1501, DukeMTMC-reID, and MSMT17, demonstrate the effectiveness of our approach.
  \end{abstract}
  

%
\IEEEpeerreviewmaketitle

\section{Introduction}

\IEEEPARstart{P}{erson} re-identification aims to match persons to a given query with visual data captured by surveillance cameras in nonoverlapping views. It has an important and wide application in video surveillance and public security. Although great advances have been witnessed in recent years, there are still many challenging issues toward its practical deployment. Due to the background clutter and the dramatic variations in viewpoints and illumination, the visual representations of the same pedestrian under different cameras may vary greatly. These factors are bottlenecks for matching pedestrians accurately across cameras. In this work, we are dedicated to learning robust and discriminative pedestrian features insensitive to backgrounds for effective person re-identification.

For person re-identification, a pedestrian is usually represented by a rectangular image patch, which inevitably contains some background regions due to the irregular shape of pedestrians.
Without precisely localizing the foreground and neglecting the background, the diverse background clutter incurs noise to the model learning and degrades the accuracy.
To alleviate the adverse influence from the backgrounds, numerous methods \cite{zhao2017spindle,wei2017glad,su2017pose,zhao2017deeply,kalayeh2018human,tian2018eliminating,song2018mask,li2018harmonious,wang2018mancs} have been proposed. In~\cite{zhao2017spindle,wei2017glad,su2017pose}, human landmark detectors were used to extract human keypoints and generate human part bounding boxes. In \cite{kalayeh2018human,song2018mask,tian2018eliminating}, segmentation models on pedestrians were applied to generate whole-body masks or multiple semantic regions. Due to the pre-trained landmark detection models and segmentation models, the body regions can be well separated from the background areas. Compared with the global features extracted from the whole images, the features from the body regions are more discriminative for person re-identification tasks without background noise. Some methods \cite{zhao2017deeply,li2018harmonious,wang2018mancs} have designed different attention models to help the networks focus on discriminative human body regions. The performance gain achieved by these methods demonstrates that removing the background influence is beneficial for person re-identification.

Although the existing methods have achieved promising results on mitigating the effects of the backgrounds, they still suffer one or more of the following limitations. 1) The human landmark detection model and segmentation model need to be pretrained with additional labeled human pose and segmentation respectively, which requires extra overhead for model training and data collection. 2) The data bias between the source datasets and the target person re-identification datasets can deteriorate the estimation of the keypoints and body masks. In particular, the existing large person re-identification datasets are usually composed of low-resolution pedestrian images, which brings remarkable challenges to adapting the pretrained models. 3) Limited by the data, the human landmark detection model and segmentation model are difficult to train together with the person re-identification model in an end-to-end manner to mutually promote each other. 4) During the inference stage, it is time consuming to generate the keypoints and body masks for individual images by these pretrained models. 5) Existing attention-based methods do not require additional training data, but the lack of strong supervision for training the networks makes them vulnerable in focusing the model attention on body regions.

To address the above issues, we propose using the camera identity information contained in the existing person re-identification datasets to separate the foreground human bodies from the background regions. Some existing methods \cite{cho2019joint,huang2016camera,lv2018unsupervised,wang2019spatial} explore the camera network topology and spatiotemporal constraints between cameras to refine the person similarities. In contrast, we exploit the camera identity information to directly train the network to learn background feature representations. By introducing the camera identity information, we can learn more discriminative person representations with the foreground branch and alleviate the negative effects of the backgrounds.

Based on the above discussion, we design an end-to-end foreground-aware network FA-Net, which aims to learn foreground-aware features effectively and efficiently. Our method contains two branches. One is the foreground feature extraction branch trained by pedestrian identity information. The other is the background feature extraction branch trained by camera identity information, which is used to constrain the target enhancement module to better distinguish the foreground from the background.
In the inference, only the foreground branch is needed to extract pedestrian features, which is very efficient. To suppress the responses in the nontarget regions, we further propose a target attention loss, which provides strong supervision for training the target enhancement module to focus on the target regions. Unlike existing works~\cite{zhao2017spindle,wei2017glad,su2017pose, kalayeh2018human,song2018mask,tian2018eliminating}, our method does not require additional human pose or segmentation but still well discriminates the foregrounds from the backgrounds with promising recognition accuracy.

In the rest of this paper, we first survey related works in Section~\ref{sec:relatedWrok}. Then, we describe our proposed framework in Section~\ref{sec:ourMethod}. After that, we evaluate our method with extensive experiments in Section~\ref{sec:experiments}. Finally, we conclude this work in Section~\ref{sec:conclusion}.

\section{Related Works}\label{sec:relatedWrok}

In this section, we first briefly introduce the progress of person re-identification. Then, we review the most related works from two aspects, \emph{i.e.,} one with a similar purpose to refine the foreground features and the other with camera evidence considered.

\subsection{Brief Overview of Person Re-identification}
Most existing person re-identification works focus on two key issues: discriminative feature representation~\cite{karanam2015person,xiong2014person} and effective distance measurement~\cite{zhao2014learning,liao2015person,yang2017person}. The background clutter, occlusion, and dramatic variations in viewpoints and pedestrian postures make it critical to extract more discriminative and robust features for person re-identification.
Given the discriminative feature representation, an effective distance metric is expected to well measure the similarities between pedestrians.

In recent years, the rapid development of convolutional neural networks (CNNs) has greatly promoted the advance of person re-identification. Based on CNNs, many discriminative feature learning methods~\cite{chen2016deep,suh2018part,sun2018beyond,li2018harmonious,zheng2019pyramidal,sun2019perceive} and distance measurement methods~\cite{shen2018end,chen2018video,shen2018person} have been proposed. Suh \emph{et al}. \cite{suh2018part} designed a network to learn a part-aligned representation for person re-identification. A two-stream network was adopted to extract appearance representations and part representations, which were further aggregated to generate part-aligned features. In \cite{sun2018beyond}, a network named part-based convolutional baseline (PCB) was proposed to extract part-level features. Shen \emph{et al}. \cite{shen2018end} proposed a Kronecker product matching module to measure the similarities of the feature maps of different persons.

\subsection{Methods Towards Refining Foreground Features}
To obtain robust representations, a key challenge is how to alleviate the influence of the backgrounds and make the network focus more on discriminative human bodies. To solve this problem, many effective methods \cite{zhao2017spindle,wei2017glad,su2017pose,zhao2017deeply,kalayeh2018human,tian2018eliminating,song2018mask,li2018harmonious,wang2018mancs} have been proposed. These methods fall mainly into the following three categories.

\textbf{Human landmark detection.} Zhao \emph{et al}. \cite{zhao2017spindle} train a model to estimate body joint locations and obtain several body subregions.
In \cite{wei2017glad}, Wei \emph{et al}. used a model pretrained on the MPII human pose dataset \cite{andriluka20142d} to estimate keypoints and crop three local body regions. In \cite{su2017pose}, a pose-driven deep convolutional (PDC) model was proposed to learn improved feature extraction and matching models, where a human pose estimation algorithm pretrained on human pose datasets was used to generate human keypoints.

\begin{figure*}[ht]
\begin{center}
\includegraphics[width=0.99\linewidth]{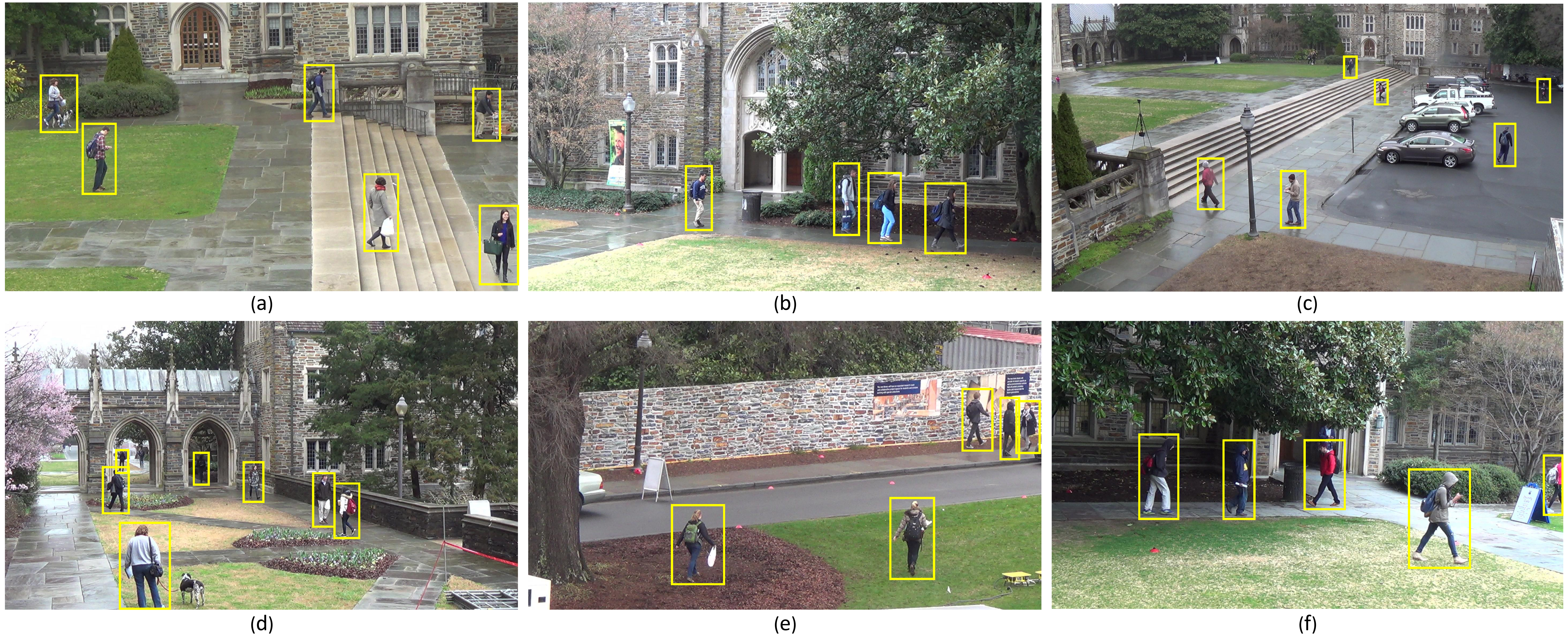}
\end{center}
   \caption{Illustration of the pedestrian images in the different scenes on DukeMTMC \cite{ristani2016performance}. Our network predicts the identities of the cameras by focusing on the background regions, which implicitly models the underlying characteristics such as viewpoints, illumination conditions and camera parameters of different scenes.}
\label{fig:scene}
\end{figure*}

\begin{figure*}[ht]
\begin{center}
\includegraphics[width=0.9\linewidth]{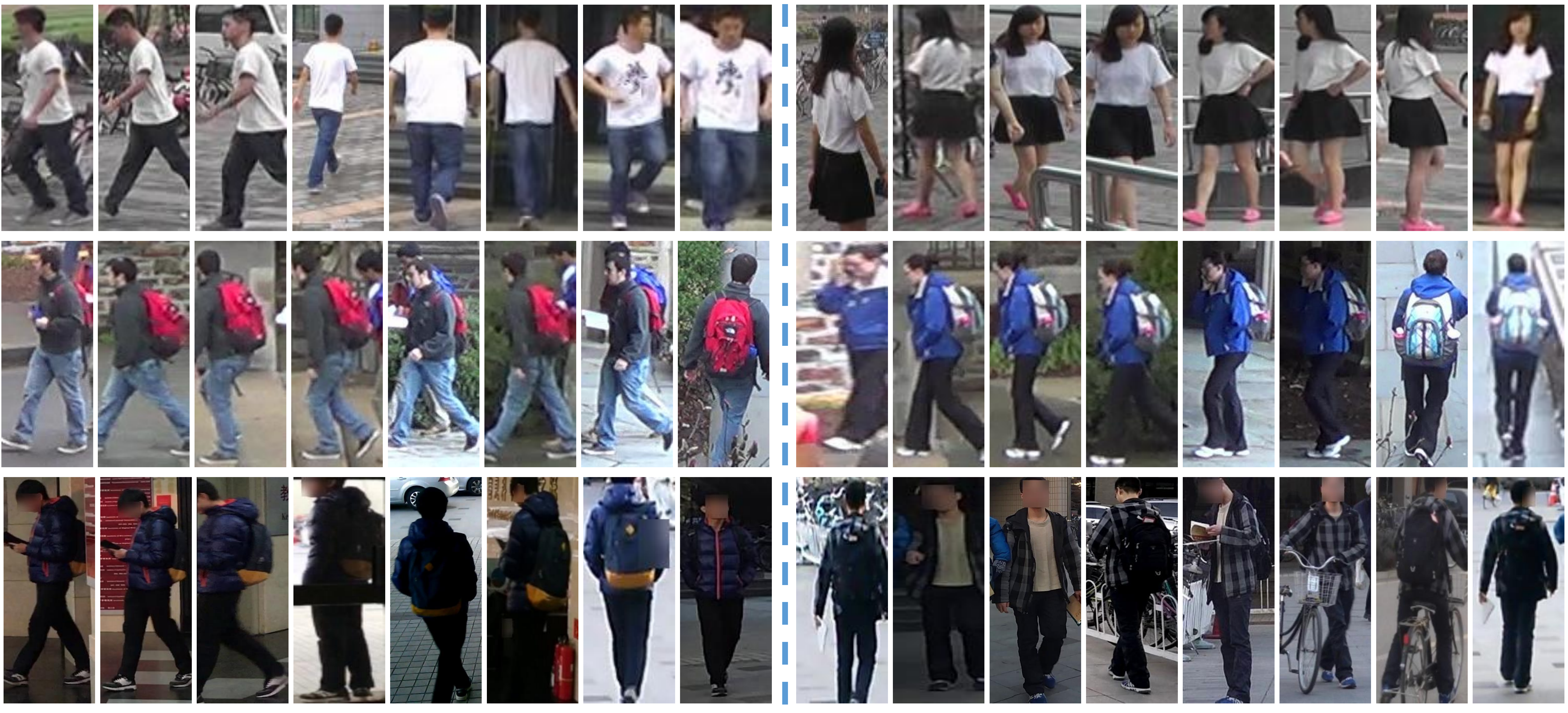}
\end{center}
   \caption{The pedestrian images on difference datasets. The three lines of images are sampled from Market-1501 \cite{zheng2015scalable}, DukeMTMC-reID \cite{zheng2017unlabeled} and MSMT17 \cite{wei2018person}. Images belonging to the same person have complex and varied backgrounds, which make it difficult to identify individuals.}
\label{fig:background}
\end{figure*}

\textbf{Segmentation-based methods.} Kalayeh \emph{et al}. \cite{kalayeh2018human} designed a SPReID model to integrate human semantic parsing in person re-identification. A human semantic parsing model was trained to segment a human body into multiple semantic regions, which were used to exploit local cues for person re-identification. Song \emph{et al}. \cite{song2018mask} used a mask-guided contrastive attention model, which extracted features separately from the body and background regions. A pretrained human segmentation model was adopted to generate a binary segmentation mask corresponding to the body and background regions. Tian \emph{et al}. \cite{tian2018eliminating} learned more discriminative person-part features based on human parsing maps generated by a person parsing network pretrained on labeled human parsing datasets.

\textbf{Attention-based methods.} Zhao \emph{et al}. \cite{zhao2017deeply} designed an attention model to generate multiple part maps. In \cite{li2018harmonious}, a harmonious attention CNN (HA-CNN) model was proposed to jointly learn the soft pixel attention and the hard regional attention along with the simultaneous optimization of feature representations. Wang \emph{et al}. \cite{wang2018mancs} proposed a fully attentional block (FAB) to localize the most discriminative local regions for person re-identification.
By applying FAB in different levels of intermediate features, they can acquire different scales of attention responses.

Different from the above works, our method aims to mitigate the influence of backgrounds more effectively and efficiently. FA-Net does not require additional human pose or segmentation datasets but still has strong supervision information to help locate the body parts and the background parts. Additionally, the background feature extraction branch is trained together with the foreground feature extraction branch. This end-to-end training strategy allows the two branches to promote each other to accurately locate the body regions and extract more robust features.

\subsection{Methods Considering Camera Information}
In addition to exploiting visual information to match pedestrians, there are some methods \cite{cho2019joint,huang2016camera,lv2018unsupervised,wang2019spatial} using the spatial context of the cameras and the temporal stamp of visual frames to constrain the learning of person similarities.
In \cite{huang2016camera,wang2019spatial}, different approaches were explored to use the spatiotemporal constraint to eliminate the irrelevant gallery images.
Lv \emph{et al}. \cite{lv2018unsupervised} proposed an unsupervised incremental learning algorithm to mine spatiotemporal patterns using the time interval of pedestrians’ transferring across different cameras. In \cite{cho2019joint}, a unified framework was designed that used the spatiotemporal relations to perform camera network topology inference.

Different from the above methods, we utilize camera information from a new perspective. Specifically, we directly use the camera identity information to guide the network to locate the background regions and help the person feature extraction model to alleviate the negative effect from the backgrounds.


\begin{figure*}[ht]
\begin{center}
\includegraphics[width=0.9\linewidth]{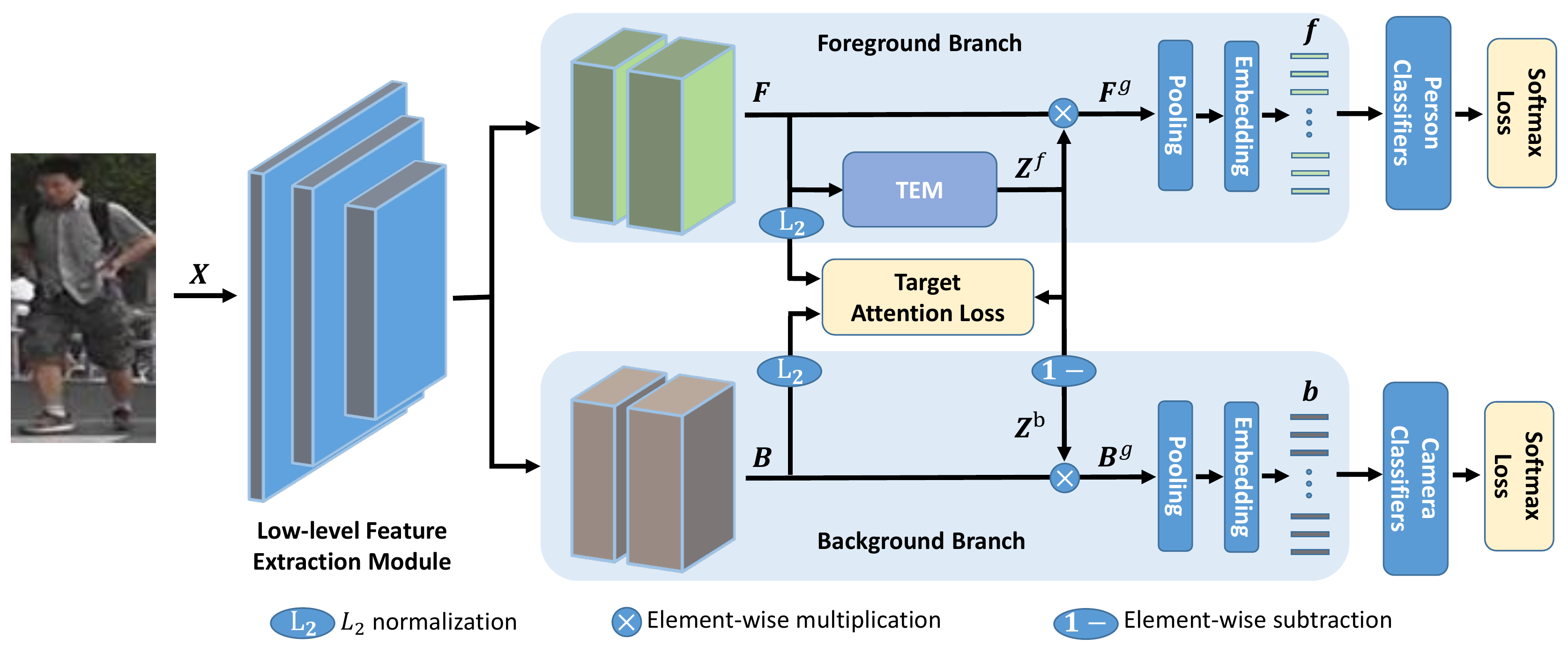}
\end{center}
   \caption{The overall architecture of the proposed method. The foreground branch and the background branch are independent of each other and do not share their weights. TEM denotes the target enhancement module. The camera classifier is trained to predict which scene the background of one image belongs to.  During the inference stage for person re-ID, the background branch is no longer needed.}
\label{fig:archi}
\end{figure*}

\section{Our Method}\label{sec:ourMethod}

In this section, we first introduce the overall architecture of the proposed end-to-end foreground-aware network for collaborative learning of foreground features and background features in Section~\ref{met:siamese}. Then, we describe our target enhancement module, target enhancement strategy and target attention loss in Section~\ref{met:tem}, Section~\ref{met:tes} and Section~\ref{met:loss}, respectively. Finally, we discuss our overall training objective in Section~\ref{met:overall}.

\subsection{Collaborative Learning of Person ID and Camera ID}\label{met:siamese}
In the video surveillance for person re-identification, each person is photographed by a certain camera as illustrated in Fig.~\ref{fig:scene}, and detected in the form of a cropped rectangular image patch, which contains not only the person as the foreground but also some portion of the scene background, as shown in Fig.~\ref{fig:background}. As a result, each person's image is characterized by two attributes, \emph{i.e.,} the person ID and the camera ID. Cropped patches belonging to the same person have similar foreground but different backgrounds, which indicates that, to identify the person ID, it is necessary to focus on the foreground and avoid the effect of the background. However, pedestrian images captured by the same camera are detected from the same scene. The foregrounds, \emph{i.e.}, the pedestrians, are usually changing, but the backgrounds are parts of the same scene and share the same camera identity. Therefore, to identify the camera identity of a pedestrian image, we should focus on the background and suppress the effects of the foreground. In summary, if a cropped person image can be decomposed into the foreground region and the background region, we can effectively learn the person ID as well as the camera ID separately.

However, in the person re-identification task, the foreground mask of a cropped person image is usually unavailable. Since the foreground exactly corresponds to the supplementary region of the background in a cropped person image, the learning of person ID and camera ID can be decoupled by introducing a pseudomask to indicate the foreground. Based on such observations, we propose a framework with two branches to mutually promote the learning of person ID and camera ID simultaneously.
As illustrated in Fig.~\ref{fig:archi}, given an input image \(\mathbf{X}\), we first extract low-level feature maps, which are then fed to two independent branches, \emph{i.e.}, the foreground branch and the background branch. The person representation and the background (\emph{i.e.,} camera) representation learned from the two branches are exploited to predict the person ID and camera ID, respectively. To facilitate  learning, we propose a new target enhancement module as well as a target attention loss, which makes two branches interact and promote each other and will be described in the next subsection.

We adopt ResNet50 \cite{he2016deep} as the backbone model, while the last global average pooling layer and the fully connected (FC) layer are removed.
The layers before the res\_conv4 block are adopted as the low-level feature extraction module. The remaining blocks of ResNet50 are copied into two independent branches, \emph{i.e.}, the foreground branch and the background branch. The two branches do not share their weights. Given the low-level features of image \(\mathbf{X}\), the foreground branch first obtains the pedestrian feature map \(\mathbf{F} \in {\mathbb{R}}^{C \times H \times W}\) from the output of the last residual block. Similarly, the background branch extracts the raw background feature map \(\mathbf{B} \in {\mathbb{R}}^{C \times H \times W}\) from the output of the low-level feature extraction module. Based on \(\mathbf{F}\), the foreground target enhancement module (TEM) generates the corresponding spatial attention map. After being enhanced by the spatial attention maps, we obtain the gated foreground feature map \(\mathbf{F}^\mathrm{g} \in {\mathbb{R}}^{C \times H \times W}\) and gated background feature map \(\mathbf{B}^\mathrm{g} \in {\mathbb{R}}^{C \times H \times W}\).

Horizontal pyramid pooling (HPP)~\cite{fu2018horizontal} successfully enhances the discriminative capabilities of various person parts. Since our network needs to accurately distinguish the foregrounds from backgrounds, the involvement of HPP can further improve the accuracy of spatial attention map prediction in local regions. Therefore, we apply HPP on both \(\mathbf{F}^\mathrm{g}\) and \(\mathbf{B}^\mathrm{g}\) to obtain features with four horizontal pyramid scales. The four scales have 1, 2, 4, and 8 spatial stripes. For each scale, feature maps are sliced to the corresponding number of stripes. The features in each stripe are then pooled and embedded into a 256-dim feature vector. Given a foreground feature vector, the corresponding person classifier predicts the person identity and calculates the softmax loss. Each background feature vector is fed to a corresponding camera classifier to predict the camera identity and calculate the softmax loss.

Although the foreground branch and the background branch share a similar architecture, they are trained with different objectives. The foreground branch predicts the person identity as the target by focusing on the foreground human body regions, while the background branch predicts the camera identity as the target by focusing on the background regions.
Ideally, there should be no overlap between the focused regions of the two branches.
In the following, we introduce how to make the two branches share their complementary knowledge with a target enhancement module to benefit each other in model training.

\begin{figure}[t]
\begin{center}
\includegraphics[width=0.9\linewidth]{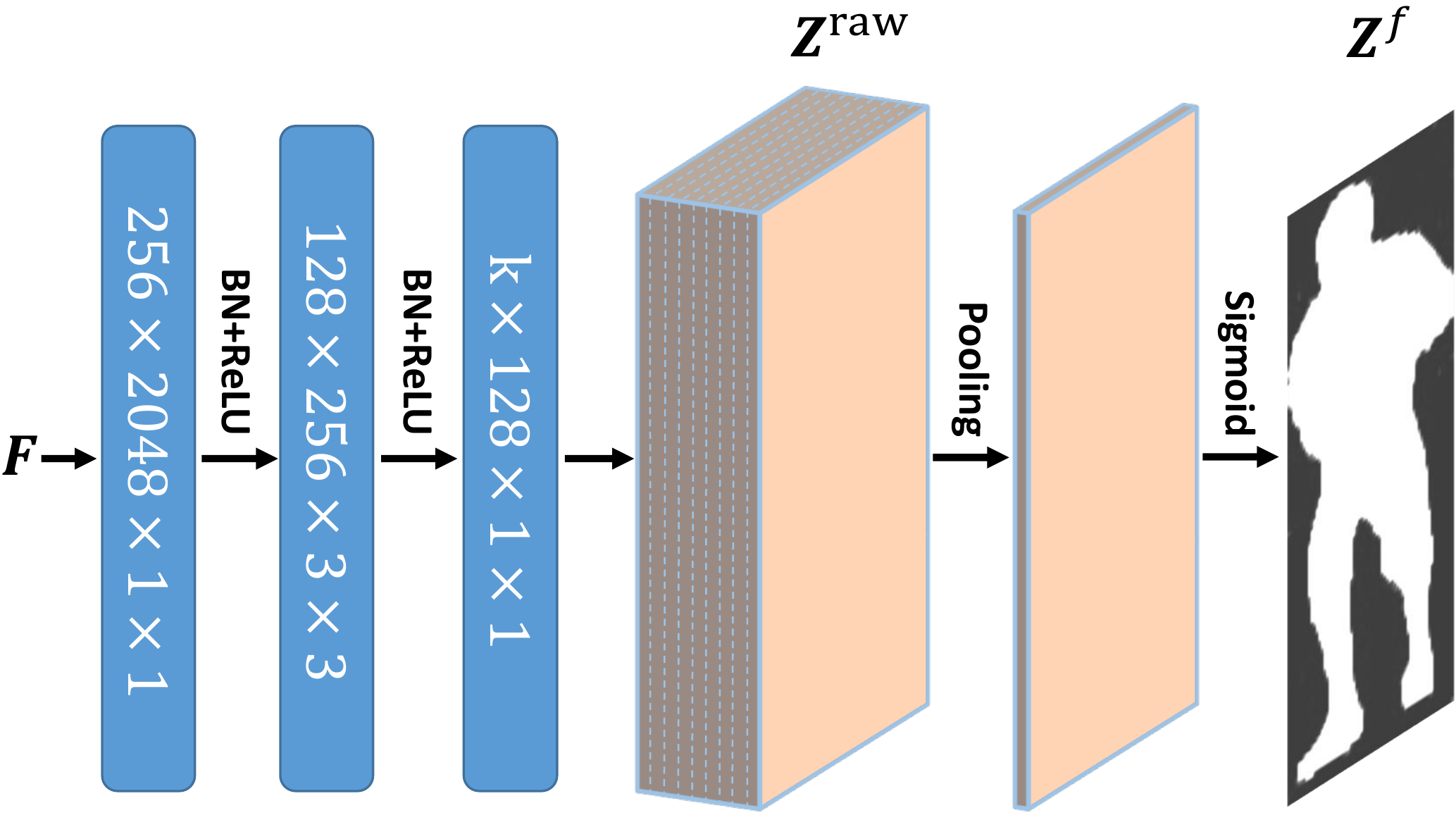}
\end{center}
   \caption{The architecture of the target enhancement module TEM.}
\label{fig:tem}
\end{figure}

\subsection{Target Enhancement Module}\label{met:tem}
The target enhancement module (TEM) aims to generate a pseudomask to indicate the target (\emph{i.e.,} foreground) region and restrain the responses in the nontarget (\emph{i.e.,} background) region, which is the key to the collaborative learning of the two branches.
The architecture of our TEM is shown in Fig.~\ref{fig:tem}. We first feed the raw feature map \(\mathbf{F} \in {\mathbb{R}}^{C \times H \times W}\) into two convolutional blocks. Each consists of three consecutive operations: a convolutional layer, a batch normalization (BN) layer and a rectified linear unit (ReLU). The first convolutional block has 256 filters. The kernel size is set to \(1 \times 1\), to reduce the feature dimension. For the second convolutional block with 128 filters, the kernel size is set to \(3 \times 3\), which increases the receptive field of this module. Then, the output of the two blocks is fed into another convolutional layer with a \(1 \times 1\) kernel size to generate the spatial attention map \(\mathbf{Z}^\mathrm{raw} \in {\mathbb{R}}^{k \times H \times W}\).  In \(\mathbf{Z}^\mathrm{raw}\), there are \(k\) channels, and each channel corresponds to a spatial attention map. We average these \(k\) spatial attention maps into spatial attention map. Finally, this spatial attention map is normalized into \([0,1]\) by the sigmoid function, which is formulated as follows:
\begin{equation}
  \mathbf{Z}^\mathrm{f} = \mathrm{sigmoid}(\frac{1}{k}\sum_{c=1}^{k}\mathbf{Z}^\mathrm{raw}(c)) \,,
\end{equation}
where \(\mathbf{Z}^\mathrm{raw}(c)\) denotes the \(c^{\mathrm{th}}\) channel of \(\mathbf{Z}^\mathrm{raw}\). \(\mathbf{Z}^\mathrm{f} \in {\mathbb{R}}^{1 \times H \times W}\) is the final foreground spatial attention map, which works as our soft foreground mask.

In some previous works~\cite{zhao2017deeply,li2018harmonious,liu2019spatial,wang2018mancs}, the spatial attention map was directly generated without channelwise pooling, which suffers an unreliability issue and limits the accuracy of the attention map.
In contrast, in the proposed TEM, we average over \(k\) spatial attention maps into the final attention map, which is more robust and accurate, as justified later in our experiments.

\subsection{Target Enhancement Strategy}\label{met:tes}
The value of each location in \(\mathbf{Z}^\mathrm{f}\) denotes the probability that the corresponding spatial location of \(\mathbf{F}\) belongs to the foreground target.
The higher probability value indicates that the TEM determines that the features in this location are more likely to belong to a body part and should be reserved, while the features in the location with lower probability are more likely to belong to the background and should be restrained. Therefore, we make use of \(\mathbf{1} -\mathbf{Z}^\mathrm{f}\) to denote the probability that the corresponding spatial location of \(\mathbf{F}\) belongs to the background target. In other words, \(\mathbf{Z}^\mathrm{b} = \mathbf{1} -\mathbf{Z}^\mathrm{f}\) can be regarded as a soft background mask.

For the raw foreground feature map \(\mathbf{F}\), since we obtain the soft foreground mask \(\mathbf{Z}^\mathrm{f}\), we can use it to enhance the foreground features. The soft foreground mask is applied to each channel of the raw foreground feature map \(\mathbf{F}\), formulated as follows:
\begin{equation}
\label{equ:gate:fzz}
  \mathbf{F}^\mathrm{g} = \mathbf{F}  \odot \mathbf{Z}^\mathrm{f} \,,
\end{equation}
where \(\odot\) denotes the elementwise multiplication with broadcasting along the channels of $\mathbf F$.
The gated person feature map \(\mathbf{F}^\mathrm{g}\) is fed to the following layers to generate the final person feature representation.

Similarly, the raw background feature map \(\mathbf{B}\) is gated by the soft background mask \(\mathbf{Z}^\mathrm{b}\), which is formulated as follows:
\begin{equation}
\label{equ:gate:bzz}
  \mathbf{B}^\mathrm{g} = \mathbf{B}  \odot \mathbf{Z}^\mathrm{b} = \mathbf{B}  \odot  \left(\mathbf{1} - \mathbf{Z}^\mathrm{f}\right)\,,
\end{equation}
where \(\mathbf{B}^\mathrm{g}\) is the gated background feature. In \(\mathbf{B}^\mathrm{g}\), the features in the background regions are enhanced, and the features in the foreground regions are restrained.

Under the definition of Eq.~\ref{equ:gate:fzz} and Eq.~\ref{equ:gate:bzz}, the soft mask generated by TEM affects both foreground features and background features. This forces TEM to more accurately distinguish the foreground from the background to help both branches focus on the target areas. More importantly, with the help of TEM, the two branches collaboratively promote each other.

\subsection{Target Attention Loss}\label{met:loss}

As discussed in Section~\ref{met:tem}, the spatial attention map \(\mathbf{Z}^\mathrm{f}\) is considered as the soft foreground mask,
while the spatial attention map \(\mathbf{1} -\mathbf{Z}^\mathrm{f}\) is considered the soft background mask. In principle, in the foreground mask, the values corresponding to the background regions are expected to be close to zero, while the values of the foreground regions are expected to be close to 1.
In addition, for the raw foreground features \(\mathbf{F}\), the responses on the nontarget background regions should be small. Similarly, for the raw background features \(\mathbf{B}\), the responses on the nontarget body regions are expected to be small. Thus, we design a target attention loss (TAL) as follows:
\begin{equation}
\label{equ:tal:fzz}
  {\cal{L}_\mathrm{t}} = \mathrm{avg} \left[ \mathbf{F}^{\ell^2} \odot  \left(\mathbf{1} - \mathbf{Z}^\mathrm{f}\right)  + \mathbf{B}^{\ell^2}  \odot  \mathbf{Z}^\mathrm{f} \right] \,,
\end{equation}
where \(\mathrm{avg}\left[ \cdot \right]\) denotes the average operation.
\(\mathbf{F}^{\ell^2}\) and \(\mathbf{B}^{\ell^2}\) are the result of performing \(\ell^2\) normalization over the spatial dimension of \(\mathbf{F}\) and \(\mathbf{B}\), which are formulated as follows,
\begin{equation}
\label{equ:tal:norm}
  \mathbf{F}^{\ell^2}(c) = \frac{\mathbf{F}(c)}{\|\mathbf{F}(c)\|_2} \,, \mathbf{B}^{\ell^2}(c) = \frac{\mathbf{B}(c)}{\|\mathbf{B}(c)\|_2} \,,
\end{equation}
where \(\mathbf{F}(c)\) and \(\mathbf{F}^{\ell^2}(c)\) correspond to the feature map of the \(c^{\mathrm{th}}\) channel of \(\mathbf{F}\) and \(\mathbf{F}^{\ell^2}\), respectively. The \(\ell^2\) normalization applied on the raw feature maps is introduced to avoid the loss simply forcing all values of features to approach zero.

Since \(\mathbf{1} - \mathbf{Z}^\mathrm{f}\) is the background target mask predicted by TEM, \( \mathbf{F}^{\ell^2} \odot  \left(\mathbf{1} - \mathbf{Z}^\mathrm{f}\right)\) denotes the response in the predicted nontarget regions of the foreground features.
Then, the minimization forces the foreground branch to focus more on the person's body and the attention map \(\mathbf{Z}^\mathrm{f}\) is required to be more accurate.
Similarly, the minimization of \( \mathbf{B}^{\ell^2}  \odot  \mathbf{Z}^\mathrm{f} \) requires the background branch to focus more on the background regions and learn better background features \(\mathbf{B}\) under the guidance of the soft mask achieved from the foreground branch. Meanwhile, this also requires TEM to distinguish well between the foreground and the background.
Therefore, the minimization of \({\cal{L}_\mathrm{t}}\) allows the two branches to promote each other, which makes better use of the opposite relationship between the two branches.

\begin{table*}[ht]
\caption{The ablation study of the proposed method on the Market-1501, DukeMTMC-reID, and MSMT17 datasets. The CMC results and mAP accuracy are reported. For the baseline network, a global average pooling is directly applied to the outputs of the modified ResNet50 backbone model to generate the final feature representations. \emph{B/L} denotes the baseline network. \emph{TEM} is the target enhancement module. \emph{BG} denotes that the background branch is added, while the soft mask generated by TEM is not applied to background features. \emph{IA} denotes that the important interaction between the two branches is adopted, \emph{i.e.}, the background features are gated by the soft mask generated by the foreground branch. \emph{TAL} denotes that the target attention loss is adopted. \emph{FA-Net} is the final architecture of our method, where the horizontal pyramid pooling HPP~\cite{fu2018horizontal} is applied. }
\label{table:ablation}
\begin{center}
\begin{tabular}{l|cccc|cccc|cccc}
    \hline
    \multirow{2}{*}{Method}  & \multicolumn{4}{c|}{Market-1501} & \multicolumn{4}{c|}{DukeMTMC-reID} & \multicolumn{4}{c}{MSMT17}\\
    \cline{2-13}
    & R1 & R5 & R10 & mAP & R1 & R5 & R10 & mAP & R1 & R5 & R10 & mAP \\
    \hline
    Baseline                                    & 89.7 & 96.4 & 97.8 & 72.7       & 77.9 & 88.9 & 91.6 & 60.1    & 64.4 & 77.4 & 81.9 & 31.4    \\
    B/L+TEM                                     & 90.9 & 96.5 & 97.7 & 73.5       & 79.5 & 88.7 & 91.7 & 60.7    & 65.4 & 78.3 & 82.8 & 33.0    \\
    B/L+TEM+BG                                  & 92.5 & 97.1 & 97.9 & 79.3       & 83.4 & 91.8 & 93.5 & 67.6    & 70.8 & 82.9 & 86.8 & 41.1    \\
    B/L+TEM+BG+IA                               & 92.9 & 97.2 & 98.2 & 79.6       & 84.3 & 92.0 & 94.6 & 67.7    & 71.6 & 83.4 & 86.8 & 41.2    \\
    B/L+TEM+BG+IA+TAL                           & 93.3 & 97.4 & 98.3 & 80.1       & 85.2 & 91.9 & 94.0 & 67.9    & 72.3 & 83.5 & 87.2 & 42.1    \\
    FA-Net (B/L+TEM+BG+IA+TAL+HPP)              & 95.0 & 97.9 & 98.6 & 84.6       & 88.7 & 93.8 & 95.5 & 77.0    & 76.8 & 86.8 & 89.8 & 51.0    \\
    \hline
  \end{tabular}
\end{center}
\end{table*}

\subsection{The Overall Training Objective}\label{met:overall}

With the proposed target attention loss \({\cal{L}_\mathrm{t}}\), the overall training objective of our approach is formulated as follows:
\begin{equation}
  {\cal{L}} = \frac{1}{2} \left( {\cal{L}_\mathrm{f}} +  {\cal{L}_\mathrm{b}} \right) + {\cal{L}_\mathrm{t}} \,,
\end{equation}
where \({\cal{L}_\mathrm{f}}\) and \({\cal{L}_\mathrm{b}}\) denote the softmax losses of the foreground branch and background branch for person-ID classification and camera-ID classification, respectively.
By minimizing \({\cal{L}}\), the proposed approach learns the foreground feature representations and the background feature representations simultaneously.
Unlike existing works \cite{zhao2017spindle,wei2017glad,su2017pose,kalayeh2018human,song2018mask,tian2018eliminating}, the prediction of the background and the training of the person re-identification model are not separate. The addition of the target enhancement module and target attention loss makes the two branches couple and promote each other, which allows our model to obtain a more accurate separation of the foreground and background. Specifically, in the the forward propagation, foreground feature extraction does not depend on the background branch. It is notable that during the inference stage for person re-ID, the background branch is no longer needed.

\section{Experiment}\label{sec:experiments}
In this section, we evaluate the proposed method on three large public image-based person re-identification datasets. We first describe the datasets and implementation details in Section~\ref{exp:datasets} and Section~\ref{exp:implement}, respectively. Then, we perform an ablation study of our method in Section~\ref{exp:ablation}. After that, we provide further analysis and discussion about FA-Net in Section~\ref{exp:further}. Finally, in Section~\ref{exp:state}, we compare our method with state-of-the-art methods.

\begin{table*}[htb]
\caption{The impact of target attention loss TAL with different setting on the performance of the proposed method. The definitions of TAL\(^{\mathrm{v1}}\) and TAL\(^{\mathrm{v2}}\) is given in Eq.~\ref{equ:tal:fz}. In the last experiments, TAL is defined as Eq.~\ref{equ:tal:fzz}.}
\label{table:ablationloss}
\begin{center}
\begin{tabular}{l|ccccc|ccccc|ccccc}
    \hline
    \multirow{2}{*}{Method}  & \multicolumn{5}{c|}{Market-1501} & \multicolumn{5}{c|}{DukeMTMC-reID} & \multicolumn{5}{c}{MSMT17}\\
    \cline{2-16}
    & R1 & R5 & R10 & R20 & mAP & R1 & R5 & R10 & R20 & mAP & R1 & R5 & R10 & R20 & mAP \\
    \hline
    B/L+TEM+BG+IA                               & 92.9 & 97.2 & 98.2 & 99.1 & 79.6       & 84.3 & 92.0 & 94.6 & 96.1 & 67.7    & 71.6 & 83.4 & 86.8 & 89.8 & 41.2    \\
    B/L+TEM+BG+IA+TAL$^{\mathrm{v1}}$             & 92.6 & 97.3 & 98.5 & 99.2 & 79.7       & 82.8 & 91.8 & 93.5 & 95.4 & 68.5    & 68.9 & 81.1 & 85.1 & 88.5 & 39.1    \\
    B/L+TEM+BG+IA+TAL$^{\mathrm{v2}}$             & 93.1 & 97.7 & 98.6 & 99.2 & 79.8           & 83.3 & 91.5 & 93.7 & 95.2 & 67.1    & 70.0 & 81.9 & 85.7 & 89.0 & 39.4    \\
    B/L+TEM+BG+IA+TAL                           & 93.3 & 97.4 & 98.3 & 99.1 & 80.1       & 85.2 & 91.9 & 94.0 & 95.6 & 67.9    & 72.3 & 83.5 & 87.2 & 90.1 & 42.1    \\

    \hline
  \end{tabular}
\end{center}
\end{table*}

\subsection{Datasets and Protocols}
\label{exp:datasets}
To evaluate our proposed methods, we select three large publicly available person re-identification datasets, namely Market-1501 \cite{zheng2015scalable}, DukeMTMC-reID \cite{zheng2017unlabeled} and MSMT17 \cite{wei2018person}. Market-1501 contains 32,668 images of 1,501 identities captured by 5 high-resolution cameras and one low-resolution camera. Images are detected by the deformable part model (DPM) \cite{felzenszwalb2010object}. The dataset is split into the training set and testing set. A total of 12,936 images of 751 identities are selected as the training set. The remaining 750 identities are used to create the gallery and query sets, which contain 19,734 and 3,368 images, respectively.

DukeMTMC-reID \cite{zheng2017unlabeled} contains the person images extracted from the DukeMTMC \cite{ristani2016performance} tracking dataset. These hand-annotated images are captured from 8 high-resolution cameras. In the standard evaluation protocol, the training set consists of 16,522 images of 702 identities. The remaining 702 identities are used as the testing set with 2,228 query images and 17,661 gallery images.
This dataset is very challenging due to the large variations within the same identity and high similarity across persons.

MSMT17 \cite{wei2018person} is a newly released large-scale person re-identification dataset that consists of 126,441 images of 4,101 identities. The images are captured by 12 outdoor cameras and 3 indoor cameras. Four days with different weather conditions in a month are selected for video collection. Videos of 3 hours each day are taken in the morning, noon and afternoon. The bounding boxes are detected by Faster RCNN \cite{ren2015faster}. In the standard evaluation protocol, 30,248 images of 1,041 identities are sampled as the training set. The remaining images of the 1,041 identities are used as the validation set. The 3,060 identities that do not appear in the training set are selected as the testing set with 11,659 query images and 82,161 gallery images.

Compared with Market-1501 and DukeMTMC-reID, MSMT17 contains more identities and images. The more camera views, both indoor and outdoor scenes and the lighting changes at different times of one day make the backgrounds more complex and challenging than previous datasets.

Following most of the previous works, we adopt the cumulated matching characteristics (CMC) table and the mean average precision (mAP) to evaluate the performance of each method. All experiments are conducted with the single query setting.
\begin{figure}[t]
\begin{center}
\includegraphics[width=0.8\linewidth]{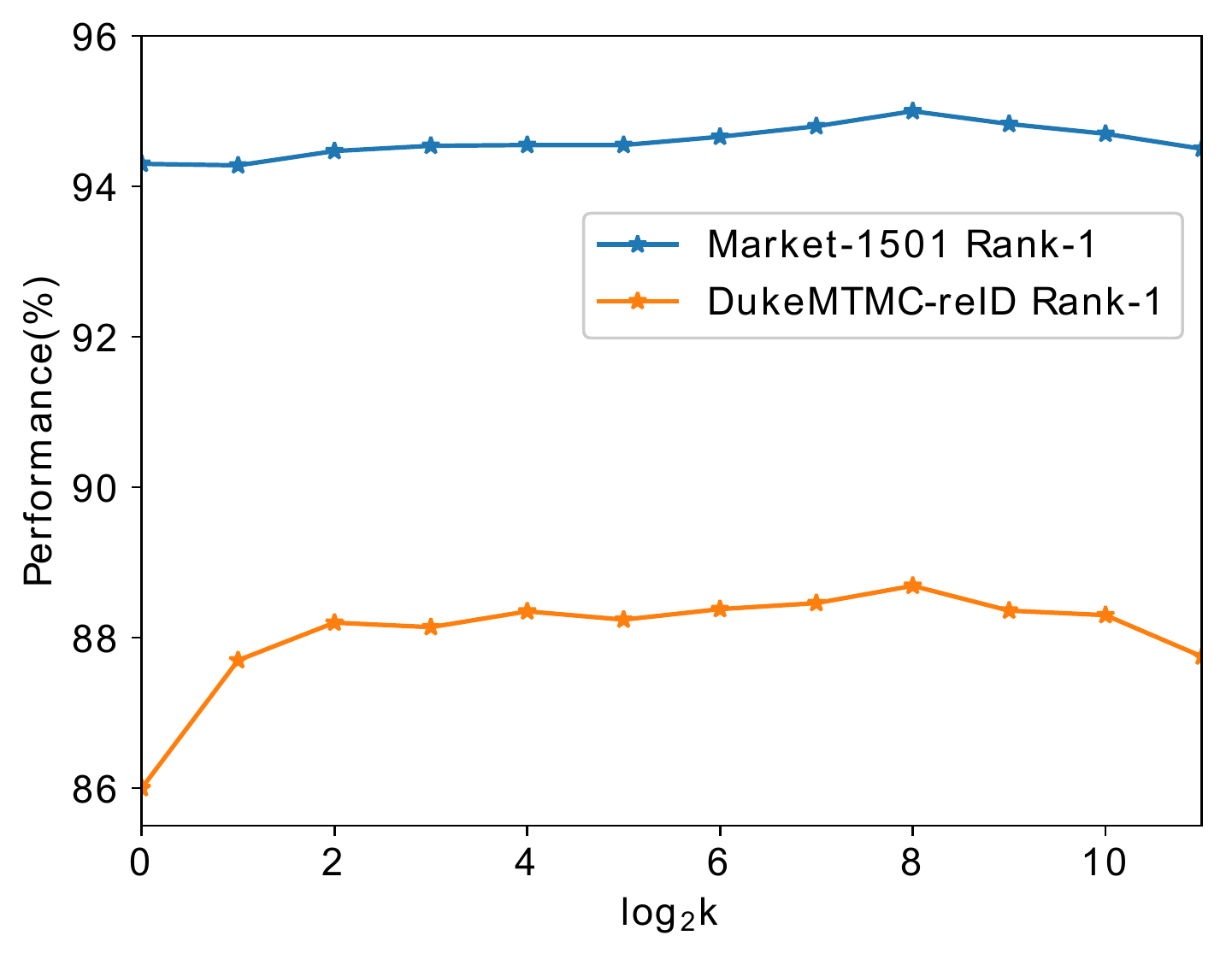}
\end{center}
   \caption{The rank-1 accuracies of FA-Net with different \(k\) values on the Market-1501 and DukeMTMC-reID datasets.}
\label{fig:alpha}
\end{figure}

\subsection{Implementation Details}
\label{exp:implement}

The backbone model ResNet50 is pretrained on the ImageNet dataset. To increase the spatial resolution, following \cite{sun2018beyond,fu2018horizontal}, the last spatial downsampling operation in the backbone network is removed. The input images of the proposed model are resized to \(384 \times 128\). Random horizontal flipping is adopted for data augmentation. In each iteration, we select images of 16 pedestrians each with 8 images as the inputs of the network in a mini-batch. The images of each pedestrian are taken from as many different cameras as possible.

The network is updated for 100 epochs by the stochastic gradient descent algorithm with a weight decay of \(5\!\times\!10^{-4}\). Following \cite{luo2019bags}, the warmup learning rate adjustment strategy is applied to bootstrap the network for better performance. The learning rate linearly increases from \(0.06\) to \(0.6\) in the first 10 epochs. Then, the learning rate is decayed to \(6\!\times\!10^{-2}\) and \(6\!\times\!10^{-3}\) at \(40^{\mathrm{th}}\) and \(80^{\mathrm{th}}\) epoch, respectively. The learning rate of the pretrained layers is set to \(0.1\!\times\) of the base learning rate. During the evaluation, the averaged feature of the original image and the horizontally flipped version is extracted for each pedestrian image. We use the cosine distance to measure the similarity of two images.

\subsection{Ablation Study}
\label{exp:ablation}

\textbf{Impact of each component.} As shown in Table~\ref{table:ablation}, we evaluate the effect of each component of our network.
The baseline network directly applies global average pooling on the outputs of the modified ResNet50 backbone model to generate the final feature representations.
After TEM is added, the network achieves 1.2\%, 1.6\% and 1.0\% improvement in the rank-1 accuracy and 0.8\%, 0.6\% and 1.6\% improvement in the mAP accuracy on Market-1501, DukeMTMC-reID and MSMT17, respectively. This indicates that TEM effectively helps the network focus more on discriminative regions.

In \emph{B/L+TEM+BG}, the background branch is added, but the background features are not gated by the soft masks.
The joint training of the two branches brings significant improvements in the rank-1 accuracies and mAP accuracies on all three datasets.
Because of the addition of the background branch, the low-level feature extraction module is shared by the two branches. To predict the camera identities, the background branch requires the low-level feature extraction module to learn additional texture and color patterns, which provides richer patterns for the extraction of foreground features.

When the main interaction between the two branches is applied, the gated features are obtained according to Eq.~\ref{equ:gate:fzz} and Eq.~\ref{equ:gate:bzz}.
The network achieves 0.4\%, 0.9\% and 0.8\% improvement in the rank-1 accuracy on Market-1501, DukeMTMC-reID and MSMT17, respectively. This is because the prediction of the soft mask generated by TEM simultaneously affects the features of both branches. To identify the camera identities of images, the network requires the soft mask to accurately distinguish between the foreground and background. The addition of new supervision information better guides the training of TEM.


After adding TAL, the performances are improved by 0.4\%, 0.9\% and 0.7\% in rank-1 accuracy and 0.5\%, 0.2\% and 0.9\% in mAP accuracy on Market-1501, DukeMTMC-reID and MSMT17, respectively. This shows that TAL helps the two branches interact better, which makes each branch focus more on its target regions and makes TEMs learn more accurate attention maps.
When HPP is adopted, another performance gain is obtained.
This is because the addition of HPP makes better use of TEM.
HPP forces the network to focus on the local regions and helps TEM improve the prediction accuracy.

\textbf{Analysis of the target enhancement module.} In TEM, we adopt the averaged results of \(k\) spatial attention maps as the final attention map.  Fig.~\ref{fig:alpha} shows the performances achieved by our method with different \(k\). We find that FA-Net achieves the best performance when \(k = 256\). When \(k = 1\), the rank-1 accuracies drop on both datasets compared to \(k = 256\). This is because directly generating the final attention map is unreliable. The spatial noise could corrupt the accuracy of the attention map. However, when the final attention map is the averaged result of several attention maps, it is more robust to noise, which is important to accurately enhance the target regions.

\begin{figure}[t]
\begin{center}
\includegraphics[width=0.8\linewidth]{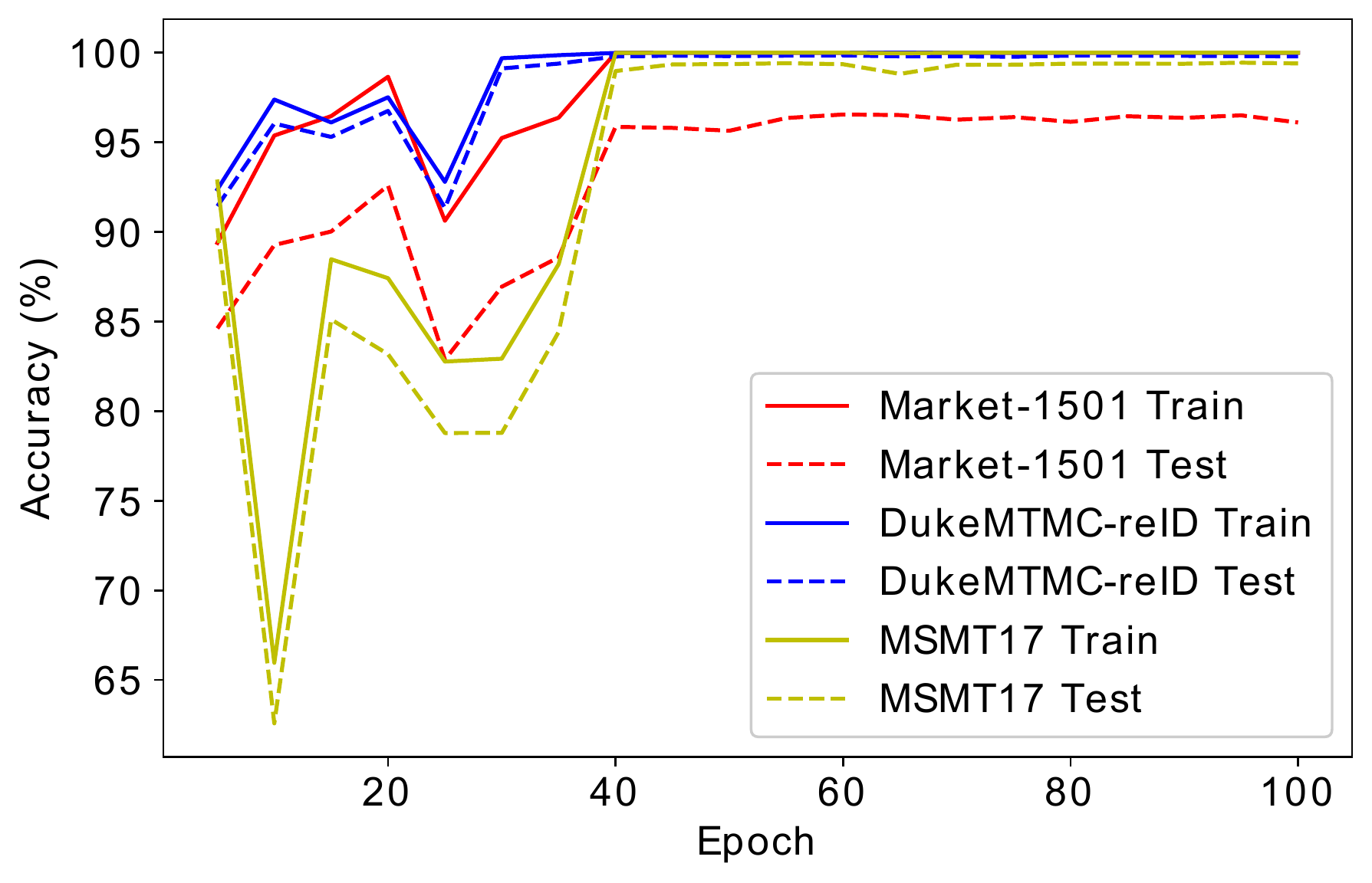}
\end{center}
   \caption{The changes in the accuracy (\%) of camera prediction on the training and test sets of three datasets. The solid and dashed lines denote the results on the training and test sets, respectively. Different colors represent different datasets. The numbers of cameras in Market-1501, DukeMTMC-reID and MSMT17 are 6, 8 and 15, respectively.}
\label{fig:acc}
\end{figure}

\begin{table}
  \caption{The accuracy (\%) of camera prediction of FA-Net on the training and test sets of three datasets.}
  \center
\begin{tabular}{l|c|c|c}

  \hline
  Dataset & Market-1501 & DukeMTMC-reID & MSMT17 \\
  \hline  
  Training Set          & 100.0 & 100.0  & 100.0  \\  \hline 
  Test Set           & 96.1 & 99.8  & 99.4  \\  \hline
\end{tabular}
\label{table:acc}
\end{table}

\begin{figure}[t]
\begin{center}
\includegraphics[width=0.99\linewidth]{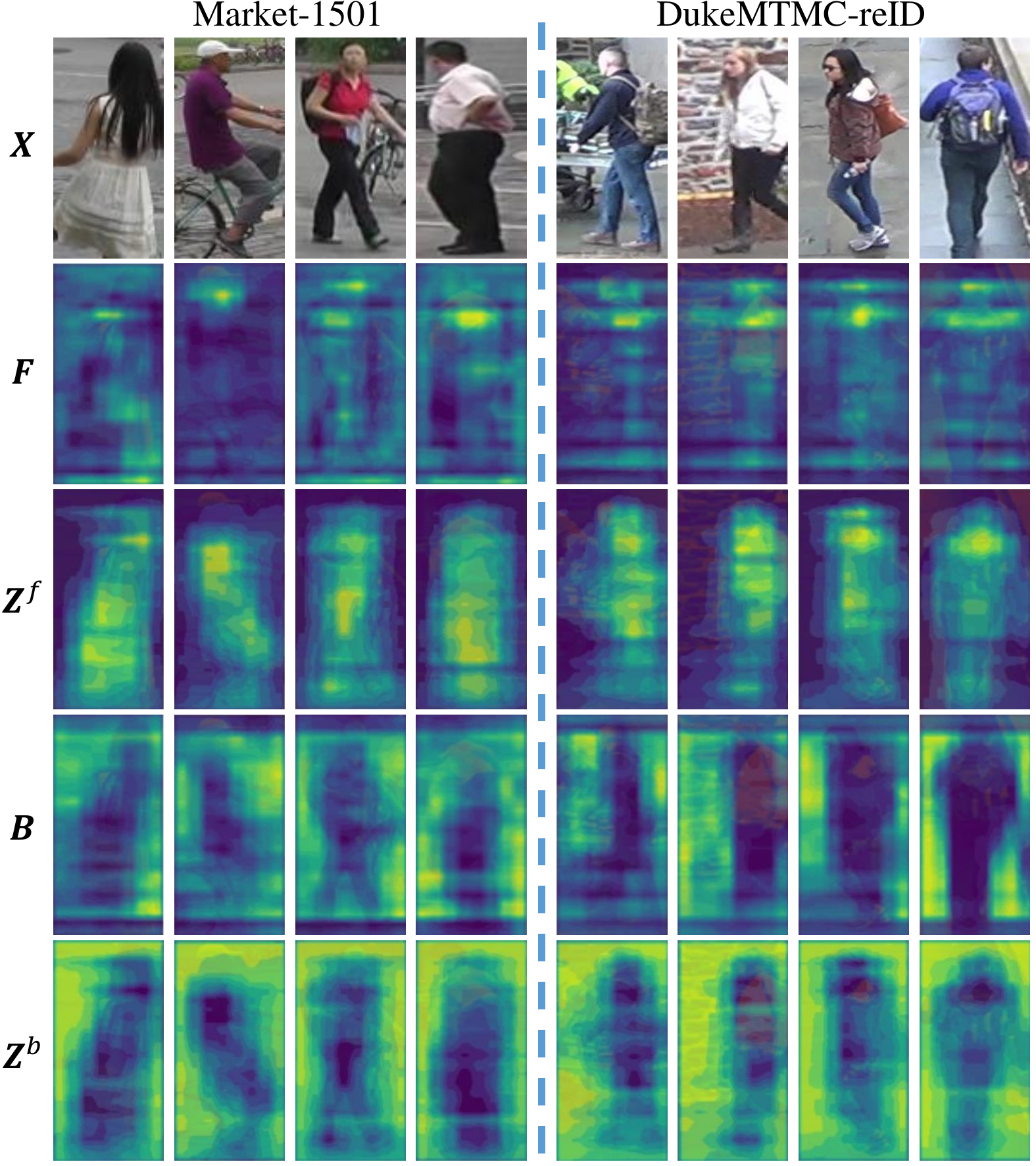}
\end{center}
   \caption{Visualization of the features and attention maps of FA-Net on the testing set.  The first row is the image input \(\mathbf{X}\). The following rows are the corresponding foreground features \(\mathbf{F}\), foreground attention map \(\mathbf{Z}^\mathrm{f}\), background features \(\mathbf{B}\) and background attention map \(\mathbf{Z}^\mathrm{b}\) of each image. The features and attention maps are displayed above the original images. The strip-shaped responses are due to the addition of horizontal pyramid pooling (HPP). }
\label{fig:feature}
\end{figure}

\textbf{Analysis of the target attention loss.}  In Table~\ref{table:ablationloss}, we show the impact of TAL with different settings on the performance of the proposed method. The other two versions of TAL are analyzed, which is formulated as follows:
\begin{equation}
\label{equ:tal:fz}
\begin{split}
  {\cal{L}^\mathrm{v1}_\mathrm{t}} &= \mathrm{avg} \left[ \mathbf{F} + \mathbf{B} \right] \,, \\
  {\cal{L}^\mathrm{v2}_\mathrm{t}} &= \mathrm{avg} \left[ \mathbf{F} \odot \left(\mathbf{1} - \mathbf{Z}^\mathrm{f}\right)  + \mathbf{B} \odot \mathbf{Z}^\mathrm{f} \right] \,,
\end{split}
\end{equation}
where \({\cal{L}^\mathrm{v1}_\mathrm{t}}\) and \({\cal{L}^\mathrm{v2}_\mathrm{t}}\) correspond to TAL\(^{\mathrm{v1}}\) and TAL\(^{\mathrm{v2}}\), respectively. In TAL\(^{\mathrm{v1}}\), the loss regularizes only the foreground and background features, which causes a slight degradation in the rank-1 accuracy of the model. This denotes that simply regularizing the features cannot boost the performance. Compared to TAL\(^{\mathrm{v1}}\), TAL\(^{\mathrm{v2}}\) achieves slight performance improvements on the rank-1 accuracy while still damaging the performance of the model on DukeMTMC-reID and MSMT17. For TAL defined in Eq.~\ref{equ:tal:fzz}, \(\ell^2\) normalization on \(\mathbf{F}\) and \(\mathbf{B}\) is applied, which boosts both the rank-1 accuracy and mAP accuracy of the model.
This is because \(\ell^2\) normalization avoids the loss simply minimizing the values of all locations. The suppression of the responses of the nontarget areas makes the model focus on the target regions and learn a better soft mask.

\subsection{Further Analysis and Discussion}
\label{exp:further}

\begin{figure*}[t]
\begin{center}
\includegraphics[width=0.99\linewidth]{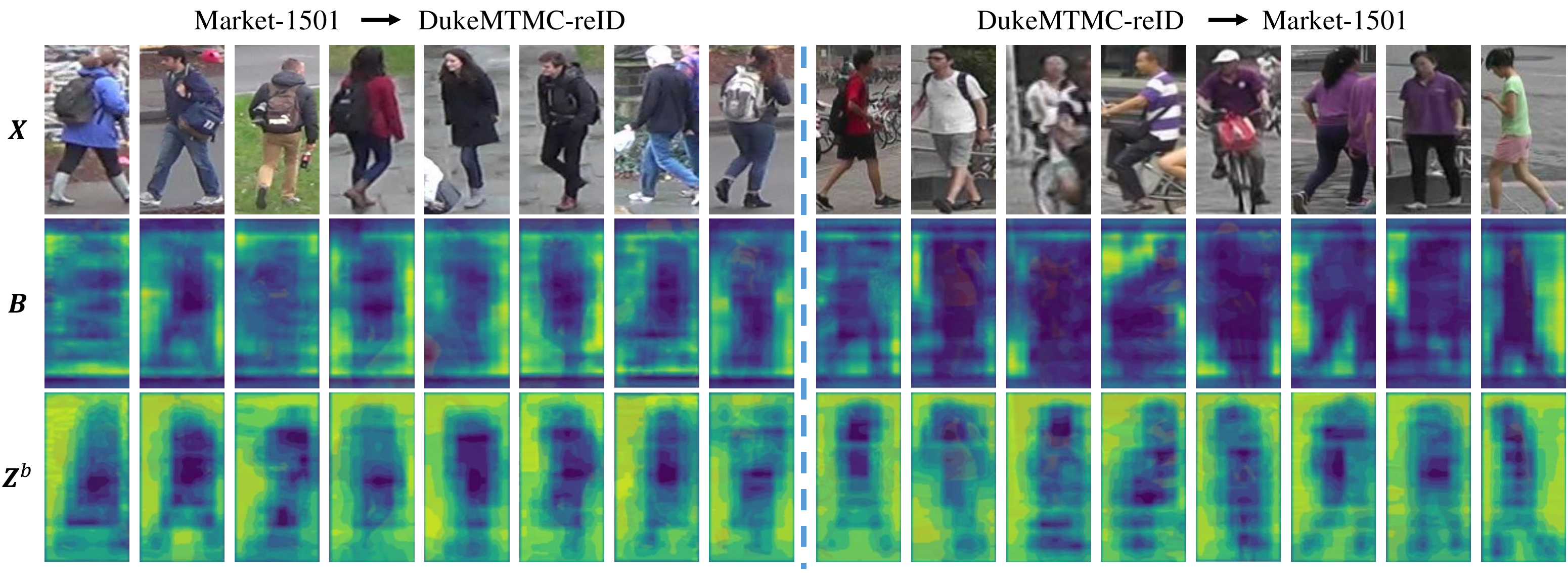}
\end{center}
   \caption{Visualization of the background features and background attention maps on the unseen scenes.  The first row is the image input \(\mathbf{X}\). The following rows are the corresponding background features \(\mathbf{B}\) and background attention map \(\mathbf{Z}^\mathrm{b}\) of each image.}
\label{fig:feanoover}
\end{figure*}

\begin{table*}[t]
\caption{The performances of our method on unseen scenes. The network is trained on one dataset and directly tested on another dataset. Therefore, there is no scene overlap between the training set and the testing set.}
\label{table:unseen}
\begin{center}
\begin{tabular}{l|ccccc|ccccc}
    \hline
    \multirow{2}{*}{Method}  & \multicolumn{5}{c|}{Market-1501 $\to$ DukeMTMC-reID} & \multicolumn{5}{c}{DukeMTMC-reID $\to$ Market-1501}\\
    \cline{2-11}
    & R1 & R5 & R10 & R20 & mAP & R1 & R5 & R10 & R20 & mAP \\
    \hline
    Baseline                                  & 28.5 & 43.5 & 50.3 & 56.8 & 13.9    & 50.4 & 67.4 & 74.2 & 80.8 & 21.5    \\
    B/L+TEM                                   & 31.8 & 46.5 & 53.1 & 59.5 & 16.1    & 52.6 & 69.7 & 76.0 & 81.9 & 23.2    \\
    B/L+TEM+BG                                & 36.9 & 52.0 & 58.3 & 65.1 & 20.3    & 54.3 & 71.0 & 77.2 & 82.7 & 24.7    \\
    B/L+TEM+BG+IA                             & 38.0 & 53.7 & 60.2 & 66.1 & 20.7    & 55.1 & 72.0 & 78.3 & 83.8 & 25.3    \\
    B/L+TEM+BG+IA+TAL                         & 39.4 & 54.9 & 60.6 & 66.0 & 21.5    & 56.2 & 73.0 & 78.5 & 83.0 & 25.5    \\
    FA-Net (B/L+TEM+BG+IA+TAL+HPP)            & 49.3 & 63.5 & 69.0 & 73.7 & 30.7    & 65.1 & 79.3 & 84.4 & 89.0 & 34.2    \\
    \hline
  \end{tabular}
\end{center}
\end{table*}

\textbf{Is it feasible to use camera identity information to guide the network to learn background features?} 
For person re-identification, due to factors such as viewpoints, illumination conditions and camera parameters, there are large gaps between the scene data captured by different cameras. Even for the same object, the features of its images captured by different cameras may vary greatly. This creates many challenges to person re-identification. However, in our method, we take advantage of this toward problem solving. Our background branch predicts the identities of the cameras by focusing on the background regions, which implicitly models the underlying characteristics such as viewpoints, illumination conditions, and camera parameters of different scenes. Even though there are large visual differences in different areas of the same scene, they share the same characteristics. For objects in different scenes, even if they look the same, those underlying characteristics can help the network to correctly determine which scenes they belong to.

As shown in Fig. \ref{fig:scene}, on the DukeMTMC-reID dataset, the lawn area and stair area of scene \emph{a} are very different, while the stone road of scene \emph{a} and the stone road of scene \emph{c} have the same material and are very similar. However, our experimental results of Table \ref{table:acc} and Fig. \ref{fig:acc} prove that our model predicts the identities of the cameras very accurately on both the training and test sets of the three datasets. These experiments fully demonstrate that our method can effectively model the underlying characteristics of different scenes and accurately predict camera identities.

We show some examples of the features and attention maps of testing images generated by FA-Net in Fig.~\ref{fig:feature}. The responses of the background features are mainly in the background regions, which means that FA-Net models the underlying characteristics of different scenes and predicts camera identity by focusing on background areas.

By observing the soft masks of the foreground and background, we see that TEM well distinguishes foregrounds and backgrounds. This benefits from the guidance from both branches and the addition of target attention loss. With the soft masks, TEM forces the two branches to focus on the target regions and learn better foreground and background representations.

\textbf{Is performance improvement due to more parameters?}
In the training stage, compared with the baseline method and \emph{B/L+TEM}, \emph{B/L+TEM+BG+IA+TAL} has more parameters due to the addition of the background branch. However, it has a similar number of parameters to the baseline network and the same number of parameters to \emph{B/L+TEM} in inference because TEM has very few parameters and the background branch is not used.
The results in Table~\ref{table:ablation} show that the performance of \emph{B/L+TEM+BG+IA+TAL} is improved significantly over the baseline model. Rank-1 accuracies are improved by \textbf{3.6\%}, \textbf{7.3\%} and \textbf{7.9\%} and mAP accuracies are improved by \textbf{7.4\%}, \textbf{7.8\%} and \textbf{10.7\%} on Market-1501, DukeMTMC-reID and MSMT17, respectively.
Compared to \emph{B/L+TEM}, \emph{B/L+TEM+BG+IA+TAL} boosts rank-1 accuracies by \textbf{2.4\%}, \textbf{5.7\%} and \textbf{6.9\%} and mAP accuracies by \textbf{6.6\%}, \textbf{7.2\%} and \textbf{9.1\%} on Market-1501, DukeMTMC-reID and MSMT17, respectively.
This means that the significant performance improvements achieved by our method are mainly due to the better feature extraction rather than more parameters. Specifically, the significant performance gains under the same number of parameters indicate that our approach is more efficient and helps reduce the computational overhead of large-scale person re-identification.

\textbf{Evaluation of unseen scenes.}
In our method, we use camera identities to guide the background branch to learn the background features and constrain the learning of the soft spatial mask to help the foreground feature focus on the person's body parts. The involvement of the background branch raises one question: can the proposed method still improve performance under unknown backgrounds?

To verify the effectiveness of the proposed method on unseen scenes, we train the network on one dataset and directly test it on another dataset. The collections of these two datasets are in different scenes and use different camera settings. Therefore, our model is tested on the new scenes, which have different backgrounds from the training set. As shown in Table~\ref{table:unseen}, we observe that even in the unseen scenes, every module of our method improves the performance. This shows that each module of FA-Net is still effective even in a new scene.  The background branch is introduced to help the low-level feature extraction module learn richer patterns and regularize the learning of TEM, and it is abandoned in inference. Therefore, FA-Net still performs well in the unseen scene.

We show some examples of the background features and background soft masks generated by our method on the unseen scenes in Fig.~\ref{fig:feanoover}. Even in unseen scenes, the background branch still focuses on the backgrounds, and the TEM well distinguishes the foregrounds and backgrounds. Specifically, for the first testing image when training on Market-1501, this kind of wall does not appear in the training dataset, but the responses in the features generated by the background branch appear at the wall.
This indicates that the background model trained using camera identity information is generalized to unseen scenes.

\textbf{Does using camera identity information require additional data collection overhead?}
In an intelligent surveillance system, after we retrieve the image of the person of interest, we usually need to further know the person's location. This is available according to the location of the camera that captures this image. This indicates that in practical applications, it is necessary to record which camera each image comes from. The recording of camera identity information is very easy and does not require manual labeling. Most of the existing person re-identification datasets also record the camera identity of each image. Instead, the methods based on the human landmark detection model and segmentation model require additional manually labeled datasets. This shows that using camera identity information is more economical and does not incur the overhead of additional data collection for many practical applications.

\subsection{Comparison with the State-of-the-Art Methods}
\label{exp:state}

As shown in Table~\ref{table:market}, we first compare our method with the related works on the Market-1501 and DukeMTMC-reID datasets. Some approaches that attempt to remove the influence from the backgrounds are included, such as the human landmark detection method GLAD~\cite{wei2017glad}, segmentation method SPReID~\cite{kalayeh2018human}, and attention-based method HA-CNN~\cite{li2018harmonious}.
Our approach achieves 95.0\% rank-1 accuracy and 84.6\% mAP accuracy on the Market-1501 dataset, 88.7\% rank-1 accuracy and 77.0\% mAP accuracy on the DukeMTMC-reID dataset.

\begin{table}
\caption{Comparison with the related methods on Market-1501 and DukeMTMC-reID. The mAP and rank-1 accuracies are reported. RK denotes the re-ranking operation \cite{zhong2017re}.}
\label{table:market}
\begin{center}
\begin{tabular}{l|c|cc|cc}
    \hline
    \multirow{2}{*}{Method} & \multirow{2}{*}{Reference} & \multicolumn{2}{c|}{Market-1501} & \multicolumn{2}{c}{DukeMTMC-reID}\\
    \cline{3-6}
    &  & R1 & mAP & R1 & mAP \\
    \hline
    CAN~\cite{liu2017end}                &  TIP'17 & 60.3  &  35.9  & -  & -  \\
    GLAD~\cite{wei2017glad}                & MM'17  & 89.9  & 73.9     & -  & -  \\
    MGCAM~\cite{song2018mask}   & CVPR'18  & 83.8  & 74.3     & -  & -  \\
    AACN~\cite{xu2018attention}                & CVPR'18  & 85.9  & 66.9     & 76.8  & 59.3  \\
    HA-CNN~\cite{li2018harmonious}                & CVPR'18  & 91.2  & 75.7     & 80.5  & 63.8  \\
    SPReID\cite{kalayeh2018human}                &  CVPR'18 & 92.5  & 81.3     & 84.4  & 71.0  \\
    FD-GAN~\cite{ge2018fd}                &  NIPS'18 & 90.5  & 77.7     & 80.0  & 64.5  \\
    PABR~\cite{suh2018part}                & ECCV'18  & 91.7  & 79.6     & 84.4  & 69.3  \\
    PCB~\cite{sun2018beyond}   & ECCV'18  & 92.3  & 77.4     & 81.7  & 66.1  \\
    PCB+RPP~\cite{sun2018beyond}   &  ECCV'18 & 93.8  & 81.6     & 83.3  & 69.2  \\
    LITM+GHIS~\cite{zhang2018learning}                & AAAI'19  & 93.9  & 83.9     & 85.9 & 74.5  \\
    HPM~\cite{fu2018horizontal}                &  AAAI'19 & 94.2  & 82.7     & 86.6  & 74.3  \\
    IANet~\cite{hou2019interaction}             & CVPR'19  & 94.4  & 83.1     & 87.1  & 73.4  \\
    \hline
    FA-Net      &  This work & \textbf{95.0}  & \textbf{84.6}     & \textbf{88.7} &  \textbf{77.0} \\
    \hline
    AACN+RK~\cite{xu2018attention}                & CVPR'18  & 88.7  & 83.0     & -  & -  \\
    PABR+RK~\cite{suh2018part}                & ECCV'18  & 93.4  & 89.9     & 88.3  & 83.9  \\
    PCB+RPP+RK~\cite{sun2018beyond}   & ECCV'18  & 95.1 & 91.9     & -  & -  \\
    FA-Net+RK      & This work  & \textbf{95.8}  & \textbf{93.4}     & \textbf{91.5}  & \textbf{88.9}  \\
    \hline
  \end{tabular}
\end{center}
\end{table}
\begin{table}
\caption{Comparison with the existing methods on MSMT17.}
\label{table:msmt17}
\begin{center}
\begin{tabular}{l|c|cccc}
    \hline
    Method & Reference & R1 & R5 & R10 & mAP \\
    \hline
    GoogLeNet \cite{szegedy2015going}    & CVPR'15  & 47.6  & 65.0  & 71.8  & 23.0  \\
    PDC \cite{su2017pose}    &  ICCV'17 & 58.0  & 73.6  & 79.4  & 29.7  \\
    GLAD \cite{wei2017glad}    &  MM'17 & 61.4  & 76.8  & 81.6  & 34.0  \\
    PCB + RPP~\cite{sun2018beyond}   &  ECCV'18 & 68.2  & 81.2     & 85.5  & 40.4  \\
    IANet~\cite{hou2019interaction}   & CVPR'19  & 75.5  & 85.5     & 88.7  & 46.8  \\
    \hline
    FA-Net      &  This work & \textbf{76.8}  & \textbf{86.8}     & \textbf{89.8}  & \textbf{51.0}  \\
    \hline
  \end{tabular}
\end{center}
\end{table}

Compared with the segmentation method SPReID~\cite{kalayeh2018human}, our method improves the rank-1 accuracy by 2.5\% and 4.3\% and mAP accuracy by 3.3\% and 6.0\% on Market-1501 and DukeMTMC-reID, respectively.
This indicates that our method mitigates the impact of the backgrounds and achieves better performance even without the additional human pose or segmentation datasets.
Compared to the attention-based method HA-CNN~\cite{li2018harmonious}, our method improves the rank-1 accuracies by 3.8\% and 8.2\% and mAP accuracies by 8.9\% and 13.2\% on Market-1501 and DukeMTMC-reID, respectively. This shows that the additional supervision information (camera identify information and TAL) effectively helps the attention module TEM to predict the target regions.
In IANet~\cite{hou2019interaction}, a spatial interaction-and-aggregation module (SIA) was proposed to deal with large variations in body pose and scale, which makes the network learn more robust foreground features.
Compared to IANet~\cite{hou2019interaction}, our method improves the rank-1 accuracies by 0.6\% and 1.6\% and mAP accuracies by 1.5\% and 3.6\% on Market-1501 and DukeMTMC-reID, respectively. This indicates that it is effective to use the complementary knowledge about foreground and background to learn foreground masks to enhance foreground features.

In Table~\ref{table:msmt17}, we compare our method with the existing methods on the MSMT17 dataset. Compared with IANet~\cite{hou2019interaction}, our method boosts the rank-1 accuracy by 1.3\% and the mAP accuracy by 4.2\%. It is worth noting that the images of MSMT17 have more complex backgrounds due to the 15 camera views with both indoor and outdoor scenes and the lighting changes at different times in one day. The significant performance improvement achieved on such a challenging dataset demonstrates the effectiveness of our method in handling the effects from the backgrounds and extracting more robust and discriminative pedestrian features.

\section{Conclusion}\label{sec:conclusion}
In this paper, we propose an end-to-end foreground-aware network for person re-identification. To alleviate the influence of the backgrounds, our method learns a soft foreground mask and locates the background regions using the camera identities available in the existing person re-identification datasets, rather than from additional human pose or segmentation datasets.
Benefiting from the target enhancement modules and the target attention loss, the foreground branch and the background branch simultaneously promote each other and learn more robust and discriminative feature representations. Extensive experiments on three large person re-identification datasets demonstrate the effectiveness of our approach.

\section*{Acknowledgment}

The work of Wengang Zhou was supported in part by the National Natural Science Foundation of China (NSFC) under Contract 61822208  and 61632019, and in part by the Youth Innovation Promotion Association CAS under Grant 2018497. The work of Houqiang Li was supported by NSFC under Contract 61836011 and 62021001. The work is supported by GPU cluster built by MCC Lab of Information Science and Technology Institution, USTC.


\bibliographystyle{IEEEtran}
\bibliography{bare_jrnl}

%

\begin{IEEEbiography}[{\includegraphics[width=1in,height=1.25in,clip,keepaspectratio]{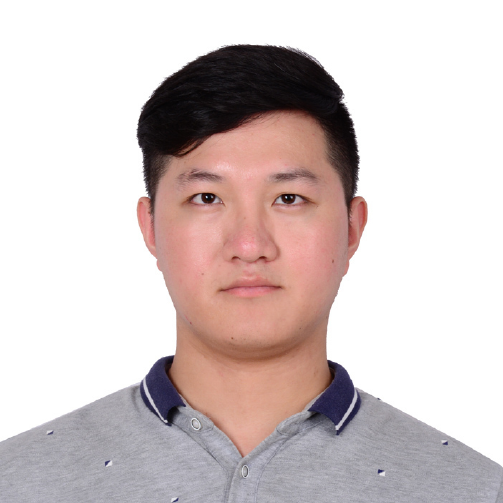}}]{Yiheng Liu}
received the B.E. degree in electronic information engineering from the University of Science and Technology of China (USTC), Hefei, China, in 2017. He is currently pursuing the Ph.D. degree in information and communication engineering with the Department of Electronic Engineering and Information Science, USTC. 

His research interests include person re-identification and computer vision.
\end{IEEEbiography}

\begin{IEEEbiography}[{\includegraphics[width=1in,height=1.25in,clip,keepaspectratio]{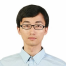}}]{Wengang Zhou}
(S'20)
received the B.E. degree in electronic information engineering from Wuhan University, China, in 2006, and the Ph.D. degree in electronic engineering and information science from University of Science and Technology of China (USTC), China, in 2011. From September 2011 to 2013, he worked as a post-doc researcher in Computer Science Department at the University of Texas at San Antonio. He is currently a Professor at the EEIS Department, USTC. 

His research interests include multimedia information retrieval and computer vision.
\end{IEEEbiography}

\begin{IEEEbiography}[{\includegraphics[width=1in,height=1.25in,clip,keepaspectratio]{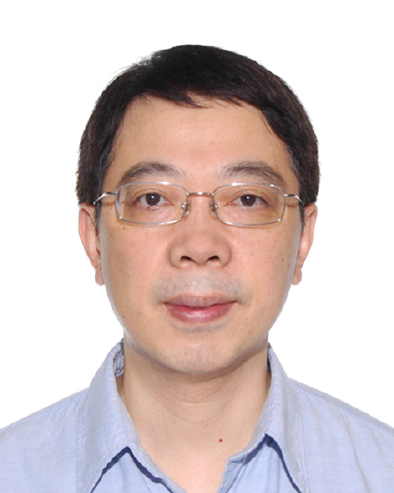}}]{Jianzhuang Liu}
(Senior Member, IEEE) received the Ph.D. degree in computer vision from The Chinese University of Hong Kong, Hong Kong, in 1997. From 1998 to 2000, he was a Research Fellow with Nanyang Technological University, Singapore. From 2000 to 2012, he was a Postdoctoral Fellow, an Assistant Professor, and an Adjunct Associate Professor with The Chinese University of Hong Kong. In 2011, he joined the Shenzhen Institutes of Advanced Technology, University of Chinese Academy of Sciences, Shenzhen, China, as a Professor. He is currently a Principal Researcher with Huawei Technologies Company Limited, Shenzhen, China. He has authored more than 150 papers. His research interests include computer vision, image processing, deep learning, and graphics.
\end{IEEEbiography}

\begin{IEEEbiography}[{\includegraphics[width=1in,height=1.25in,clip,keepaspectratio]{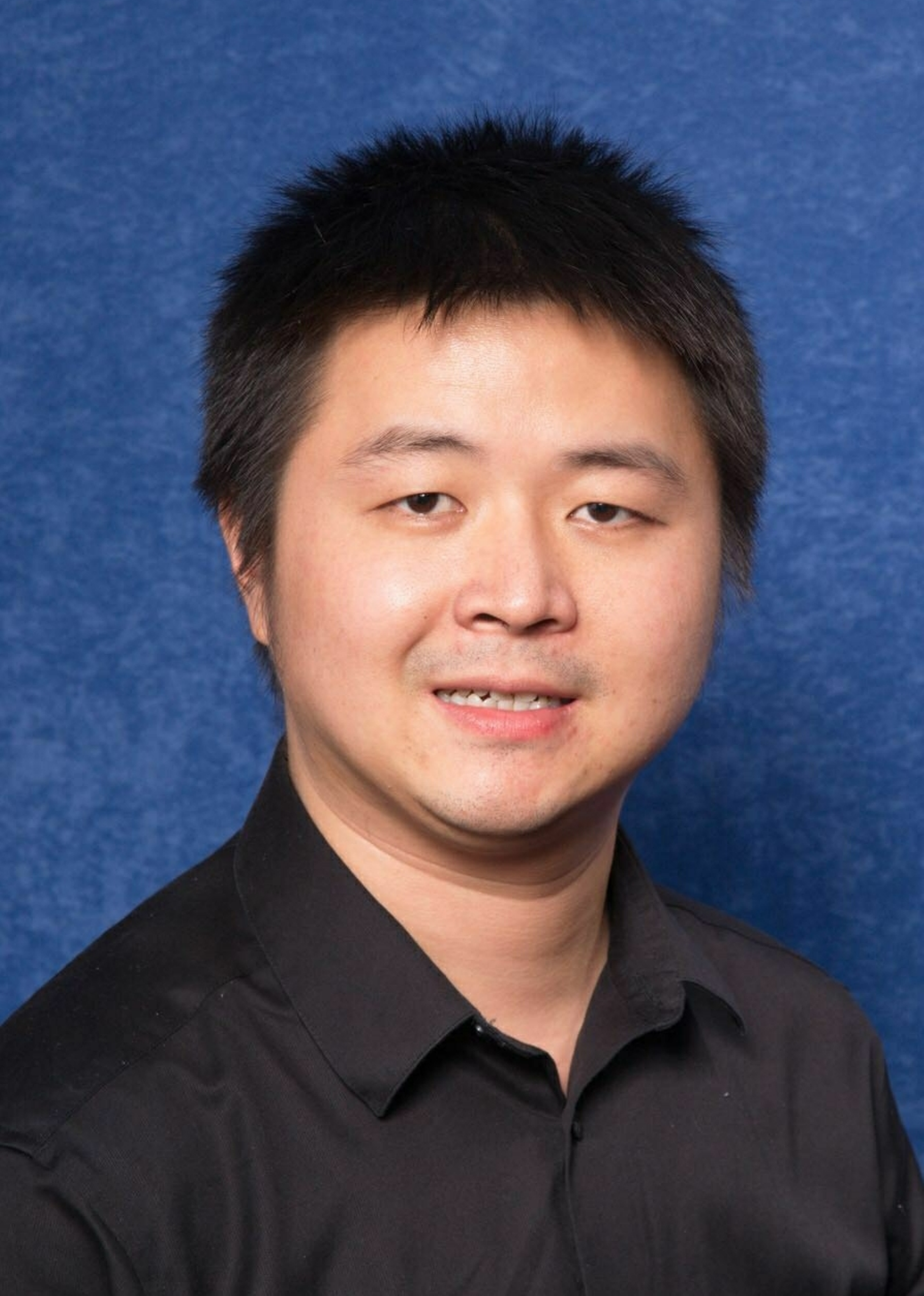}}]{Guo-Jun Qi}
(M'14-S'18) is the Chief Scientist leading and overseeing an international R\&D team in the domain of multiple intelligent cloud services, including smart cities, visual computing service, medical intelligent service, and connected vehicle service at Futurewei, since August 2018. He was a faculty member in the Department of Computer Science and the director of MAchine Perception and LEarning (MAPLE) Lab at the University of Central Florida since August 2014. Prior to that, he was also a Research Staff Member at IBM T.J. Watson Research Center, Yorktown Heights, NY. 

His research interests include machine learning and knowledge discovery from multi-modal data sources (e.g., images, videos, texts, and sensors) in order to build smart and reliable information and decision-making systems. His research has been sponsored by grants and projects from government agencies and industry collaborators, including NSF, IARPA, Microsoft, IBM, and Adobe. 

Dr. Qi has published over 150 papers in a broad range of venues, such as Proceedings of IEEE, IEEE T PAMI, IEEE T KDE, IEEE T Image Processing, ICML, NIPS, CVPR, ECCV, ACM MM, SIGKDD, WWW, ICDM, SDM, ICDE and AAAI. Among them are the best student paper of ICDM 2014, “the best ICDE 2013 paper” by IEEE Transactions on Knowledge and Data Engineering, as well as the best paper (finalist) of ACM Multimedia 2007 (2015).
\end{IEEEbiography}
  
\begin{IEEEbiography}[{\includegraphics[width=1in,height=1.25in,clip,keepaspectratio]{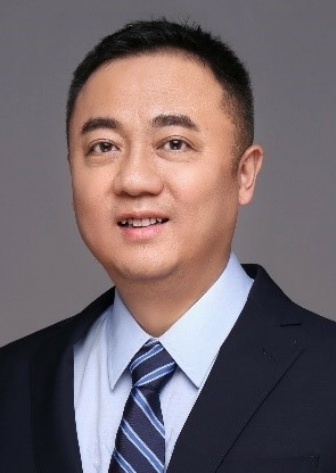}}]{Qi Tian}
(F'16) is currently a Chief Scientist in Artificial Intelligence at Cloud BU, Huawei. From 2018-2020, he was the Chief Scientist in Computer Vision at Huawei Noah’s Ark Lab. He was also a Full Professor in the Department of Computer Science, the University of Texas at San Antonio (UTSA) from 2002 to 2019. During 2008-2009, he took one-year Faculty Leave at Microsoft Research Asia (MSRA).

Dr. Tian received his Ph.D. in ECE from University of Illinois at Urbana-Champaign (UIUC) and received his B.E. in Electronic Engineering from Tsinghua University and M.S. in ECE from Drexel University, respectively. Dr. Tian’s research interests include computer vision, multimedia information retrieval and machine learning and published 590+ refereed journal and conference papers. His Google citation is over 23000+ with H-index 74. He was the co-author of best papers including IEEE ICME 2019, ACM CIKM 2018, ACM ICMR 2015, PCM 2013, MMM 2013, ACM ICIMCS 2012, a Top 10.

Dr. Tian research projects are funded by ARO, NSF, DHS, Google, FXPAL, NEC, SALSI, CIAS, Akiira Media Systems, HP, Blippar and UTSA. He received 2017 UTSA President’s Distinguished Award for Research Achievement, 2016 UTSA Innovation Award, 2014 Research Achievement Awards from College of Science, UTSA, 2010 Google Faculty Award, and 2010 ACM Service Award. He is the associate editor of IEEE TMM, IEEE TCSVT, ACM TOMM, MMSJ, and in the Editorial Board of Journal of Multimedia (JMM) and Journal of MVA. Dr. Tian is the Guest Editor of IEEE TMM, Journal of CVIU, etc. Dr. Tian is a Fellow of IEEE.
\end{IEEEbiography}

\begin{IEEEbiography}[{\includegraphics[width=1in,height=1.25in,clip,keepaspectratio]{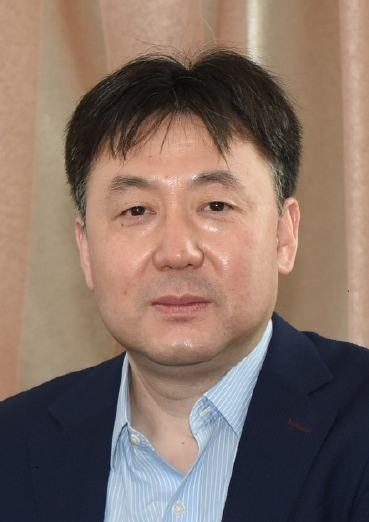}}]{Houqiang Li}
(F'21) received the B.S., M.Eng., and Ph.D. degrees in electronic engineering from the University of Science and Technology of China, Hefei, China, in 1992, 1997, and 2000, respectively, where he is currently a Professor with the Department of Electronic Engineering and Information Science. 

His research interests include multimedia search, image/video analysis, video coding and communication. He has authored and co-authored over 200 papers in journals and conferences. He is the winner of National Science Funds (NSFC) for Distinguished Young Scientists, the Distinguished Professor of Changjiang Scholars Program of China, and the Leading Scientist of Ten Thousand Talent Program of China. He served as an Associate Editor of the IEEE TRANSACTIONS ON CIRCUITS AND SYSTEMS FOR VIDEO TECHNOLOGY from 2010 to 2013. He served as the TPC Co-Chair of VCIP 2010, and he will serve as the General Co-Chair of ICME 2021. He is the recipient of National Technological Invention Award of China (second class) in 2019 and the recipient of National Natural Science Award of China (second class) in 2015. He was the recipient of the Best Paper Award for VCIP 2012, the recipient of the Best Paper Award for ICIMCS 2012, and the recipient of the Best Paper Award for ACM MUM in 2011. Dr. Li is a Fellow of IEEE.
\end{IEEEbiography}





\end{document}